\documentclass[letterpaper, 10 pt, conference]{ieeeconf}

\IEEEoverridecommandlockouts
\usepackage{cite}
\usepackage{amsmath,amssymb,amsfonts}
\usepackage{algorithmic}
\usepackage{graphicx}
\usepackage{textcomp}
\usepackage{xcolor}
\usepackage{algorithm}
\usepackage{url}
\usepackage{tabularx}
\usepackage{booktabs}
\usepackage{makecell}

\usepackage{multirow} % for non-symmetric looking tables

\usepackage{placeins}

\newcommand{\changed}[1]{{\color{red}{#1}}}

\def\BibTeX{{\rm B\kern-.05em{\sc i\kern-.025em b}\kern-.08em
    T\kern-.1667em\lower.7ex\hbox{E}\kern-.125emX}}
\begin{document}

\title{
\textbf{Occlusion Reasoning for Skeleton Extraction\\of Self-Occluded Tree Canopies}\\
%Skeletonizing Self-Occluded Tree Canopies \\in Industrial Orchards\\
%Skeleton Extraction from Self-Occluded Tree Canopies 
\thanks{C. H. Kim and G. Kantor are with Carnegie Mellon University, Pittsburgh PA, USA \texttt{\{chunghek, kantor\}@andrew.cmu.edu}}
}

\author{Chung Hee Kim and George Kantor}

\maketitle

\begin{abstract}
%A major challenge towards agricultural automation is the capability of sensing and understanding the complex task environment prior to crop interaction. 
In this work, we present a method to extract the skeleton of a self-occluded tree canopy by estimating the unobserved structures of the tree. A tree skeleton compactly describes the topological structure and contains useful information such as branch geometry, positions and hierarchy. This can be critical to planning contact interactions for agricultural manipulation, yet is difficult to gain due to occlusion by leaves, fruits and other branches. Our method uses an instance segmentation network to detect visible trunk, branches, and twigs. Then, based on the observed tree structures, we build a custom 3D likelihood map in the form of an occupancy grid to hypothesize on the presence of occluded skeletons through a series of minimum cost path searches. %The effectiveness of our approach is demonstrated through a set of experiments on a synthetic tree dataset as well as a real tree dataset collected from the field. 
We show that our method outperforms baseline methods in highly occluded scenes, demonstrated through a set of experiments on a synthetic tree dataset. Qualitative results are also presented on a real tree dataset collected from the field.

\end{abstract}

%\begin{IEEEkeywords}
%component, formatting, style, styling, insert
%\end{IEEEkeywords}

\section{Introduction}

With increasing global population and labor shortages, modern agriculture is adopting new technologies to enhance sustainability and profitability. There is growing effort in developing robotic solutions to automate repetitive and laborious tasks that often require complex and delicate interaction with crops. Harvesting and pruning, for example, may require the agent to manipulate crops by pushing aside leaves or branches to reach occluded regions before picking or cutting. Understanding the unstructured and cluttered task environment is critical to automating such challenging tasks. 

In this work, we address the perception problem, where robots must be able to sense and understand the complex task environment prior to crop interaction. We are particularly interested in extracting the skeleton of a self-occluded tree canopy by estimating the unobserved parts of the tree. A tree skeleton is useful as it describes the topological structure and contains useful information such as branch dimensions, positions, and hierarchy. This knowledge is helpful in phenotyping crop characteristics for growth assessment and can be critical for planning contact interactions, such as determining optimal pruning locations or generating trajectories to pick fruits by pulling a branch into its workspace. 

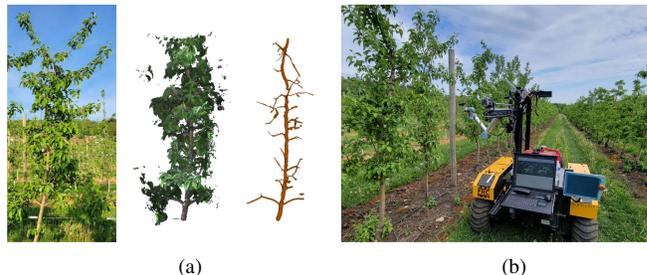
\begin{figure}[h]
\centering 
\includegraphics[width=\columnwidth]{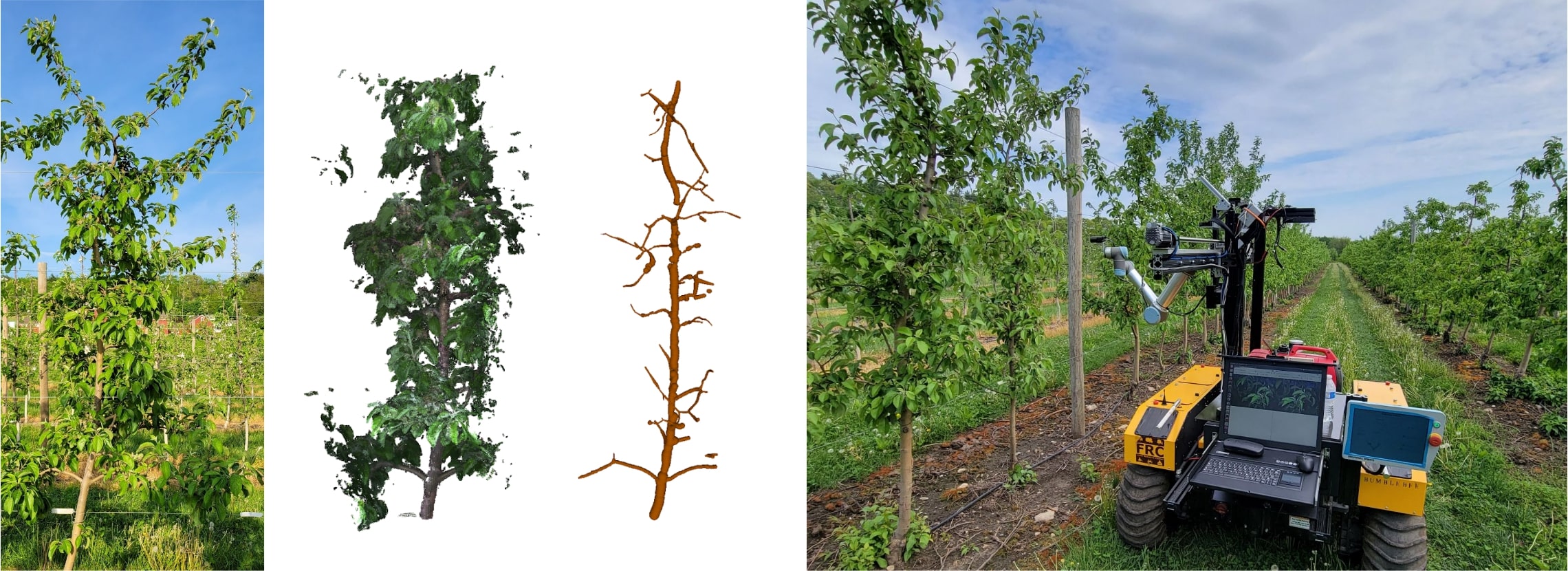}
\vspace*{-3mm}
\setlength{\unitlength}{1cm}
\begin{picture}(0,0)
\put(-2.1, 0.0){\footnotesize (a)} 
\put(2.2,  0.0){\footnotesize (b)} 
\end{picture}
\caption{(a) The 3D reconstruction (middle) and skeletonization (right) of a heavily occluded tree canopy (left). (b) Tree crops are tightly organized in rows in industrial orchard settings.}
\label{fig:intro} 
\vspace{-13pt}
\end{figure}

One of the main challenges faced during the extraction of the tree skeleton from sensor data (typically in the form of point clouds obtained from laser scanners or depth cameras) stems from noise and occlusion, which is exacerbated by the presence of foliage as depicted in Fig.~\ref{fig:intro}(a).
%As such, we propose a system to extract the skeleton of a tree canopy that is self-occluded by its own leaves, fruits or branches.  
Although there exists prior works that address skeletonization of occluded tree canopies, they are tailored towards sparse point clouds obtained from terrestrial laser scanners (TLS) and can be ineffective when applied to dense point clouds acquired from RGB-D or stereo cameras. Furthermore, it is generally assumed in prior works that the tree point cloud to be skeletonized is pre-registered from a 360 degree scan, which is difficult to obtain in industrial orchard settings where trees are tightly organized in rows as shown in Fig.~\ref{fig:intro}(b).
%scanning from all-rounded viewpoints. 

We present a novel skeletonzation method that extracts tree skeletons from one-sided views of the tree canopy by  using a depth camera attached to a robot arm. %The one-sided constraint is more practical in industrial orchard settings where trees are tightly organized in rows as shown in Fig.~\ref{fig:intro}(b). 
Our method particularly addresses the challenge of skeletonizing heavily occluded tree canopies by observing that branches in nature generally extend linearly. We use this as a heuristic assumption to approximate a 3D likelihood map in the form of an occupancy grid to predict presence of occluded branch structures by searching for minimum cost paths. We validate the effectiveness of our approach by presenting quantitative assessment on a synthetic tree dataset with known ground truth, and present qualitative results from a real tree dataset collected at an apple orchard.    

\begin{figure*}[h!]
\centering 
\includegraphics[width=\textwidth]{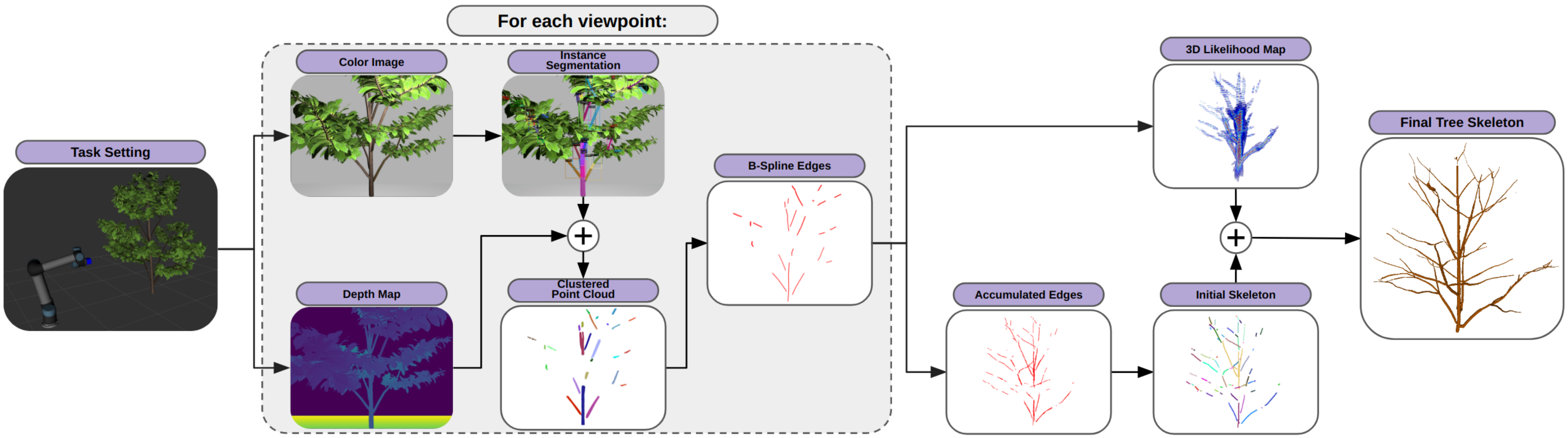}
\vspace*{-7mm}
\setlength{\unitlength}{1cm}
\begin{picture}(0,0)
\end{picture}
\caption{Our system pipeline for tree skeletonization. It takes as input a series of RGB-D images of a self-occluded tree from multiple viewpoints and outputs the underlying tree skeleton.}
\label{fig:system_overview} 
\vspace{-13pt}
\end{figure*}

\section{Related Work}
\label{sec:related_work}

A skeleton can be defined as a curve expressing the shape of an object that is consistent with the topological structure as well as the connectivity of the original object shape. Many methods have been proposed to extract skeletons from unordered 3D point clouds. A skeleton can be extracted by measuring the L1-median to determine local centers of the point cloud \cite{huang2013l1}. A Laplacian-based contraction method is proposed in \cite{cao2010point}, which collapses the input point cloud into a minimal volume using an iterative Laplacian smoothing process. 
%A mass-driven curve skeleton is extracted from point clouds by minimizing the Wasserstein distance between the mass distribution of the curve skeleton and the raw point cloud \cite{qin2019mass}. 
In \cite{tagliasacchi2009curve}, a skeleton is extracted by defining a new feature representation called rotational symmetry axis. The aforementioned methods work well with limited noise and occlusion, while we are particularly interested in skeletonizing point cloud of botanical trees which are often noisy and occluded. 

There is particular interest in extracting the skeleton of a tree canopy due to implications in automating agricultural tasks \cite{yandun2020visual} such as pruning \cite{you2022precision}, harvesting \cite{sepulveda2020robotic}, or phenotyping to estimate crop characteristics for growth assessment \cite{zhang2020apple}. Prior knowledge about tree structures are commonly used to extract tree skeletons, such as branching properties \cite{ZHANG201649}, cylindrical shape priors \cite{fu2020tree} and upright offshoots \cite{you2022semantics}. Tree skeletons are commonly represented as graphs composed of vertices and edges; \cite{delagrange2014pypetree} and \cite{bucksch2010skeltre} uses graph-based methods to extract tree skeletons. A geometry-based method is used to fit cylinders on the point cloud in a hierarchical data structure to encapsulate parent-child relations of branches \cite{hackenberg2014highly}. %These methods are tailored for trees without leaves, while we interested in skeletonizing trees with leaves which is much more challenging due to high level of occlusion.    

While the above mentioned tree skeletonization methods are tailored for trees without leaves, we address a more challenging problem in which tree canopies are self-occluded by leaves, fruits, or other branches. To address the problem of missing data, \cite{mei20173d} proposes an iterative data-completion method to recover data for 3D tree modelling. 
Tree point clouds with leaves are often first separated into branches and leaves before skeleton extraction \cite{gao2019force}, while \cite{livny2010automatic} directly generates visually convincing tree skeletons based on optimization driven by biological priors. Several works improve skeleton connectivity by joining disconnected skeletal structures through geometry-based methods \cite{sun2022fast, bremer2013derivation}; we use it as a baseline method and quantitatively show that our method outperforms \cite{sun2022fast} in occluded scenes. Despite addressing the issue of occlusion, these methods are built for sparse TLS point clouds and are inapplicable to dense point clouds. More importantly, it is assumed that the tree point cloud to be skeletonized is pre-registered from a 360 degree scan, which may be unpractical in industrial orchard setting where trees are organized in rows. %Our method extracts a skeleton by collecting images from one side of the tree which can be more practical in industrial orchard settings where trees are organized in rows.
The key contributions of this paper are:
\begin{itemize}
    \item A novel skeletonzation method that extracts tree skeletons from one-sided views of the tree canopy;
    \item A method to reason about the unobserved structures of the tree to predict the presence of occluded skeletons;
    \item Experimental results on a synthetic dataset quantitatively compared against existing baselines, as well as qualitative results on a real tree dataset collected from the field.
\end{itemize}

\section{Methodology}

\label{sec:methodology}

\subsection{System Overview}

Our tree skeletonization method aims to improve the knowledge of occluded regions in the tree that is often self-occluded by foliage, fruits, or other branches. 
%To do so, we observe that tree branches in nature generally grow straight.
Driven by a heuristic assumption on tree structures, we approximate a 3D likelihood map in the form of an occupancy grid that stores information of visible as well as occluded structures of a tree canopy to extract its skeleton.
%Our tree skeletonization method aims to improve the knowledge of occluded regions in the tree that is often self-occluded by foliage, fruits, or other branches. To do so, we observe that tree branches in nature generally grow straight. We manifest this observation into a custom 3D likelihood map in the form of an occupancy grid which stores information of visible as well as occluded structures of a tree canopy to extract its skeleton.     
Fig.~\ref{fig:system_overview} shows an overview of our framework. The system takes as input a series of color and depth images from varying viewpoints obtained from a camera attached to a robot arm with known poses. The images are passed onto an instance segmentation network to generate a semantic point cloud of branch clusters (Sec.~\ref{subsec:instance_segmentation}). 
The occupancy probability distribution of the 3D likelihood map is updated based on the observed branch clusters over a sequence of images (Sec.~\ref{subsec:likelihood_map}).
%\todo{Multivariate Gaussian distributions, fitted on each branch cluster, are sampled to create a 3D likelihood map in the form of an octree. The likelihood map obtained from each image is used to update a joint likelihood map over a series of images (Sec.~\ref{subsec:likelihood_map})}. 
The final skeleton is generated by joining disconnected branch skeletons through a series of minimum cost path searches in the likelihood map (Sec.~\ref{subsec:skeletonization}).

\subsection{Semantic Point Cloud Acquisition}
\label{subsec:instance_segmentation}

We first acquire a 3D point cloud semantically labeled with visible parts of the tree trunk, branches, and twigs (collectively referred to as branches from this point on) that are not occluded. This is achieved by projecting branch segmentation masks in 2D color images on to the 3D point cloud obtained from the depth image. The points corresponding to branches in the semantic point cloud are clustered based on the projected masks.

\begin{figure}[h!]
\centering 
\includegraphics[width=0.8\columnwidth]{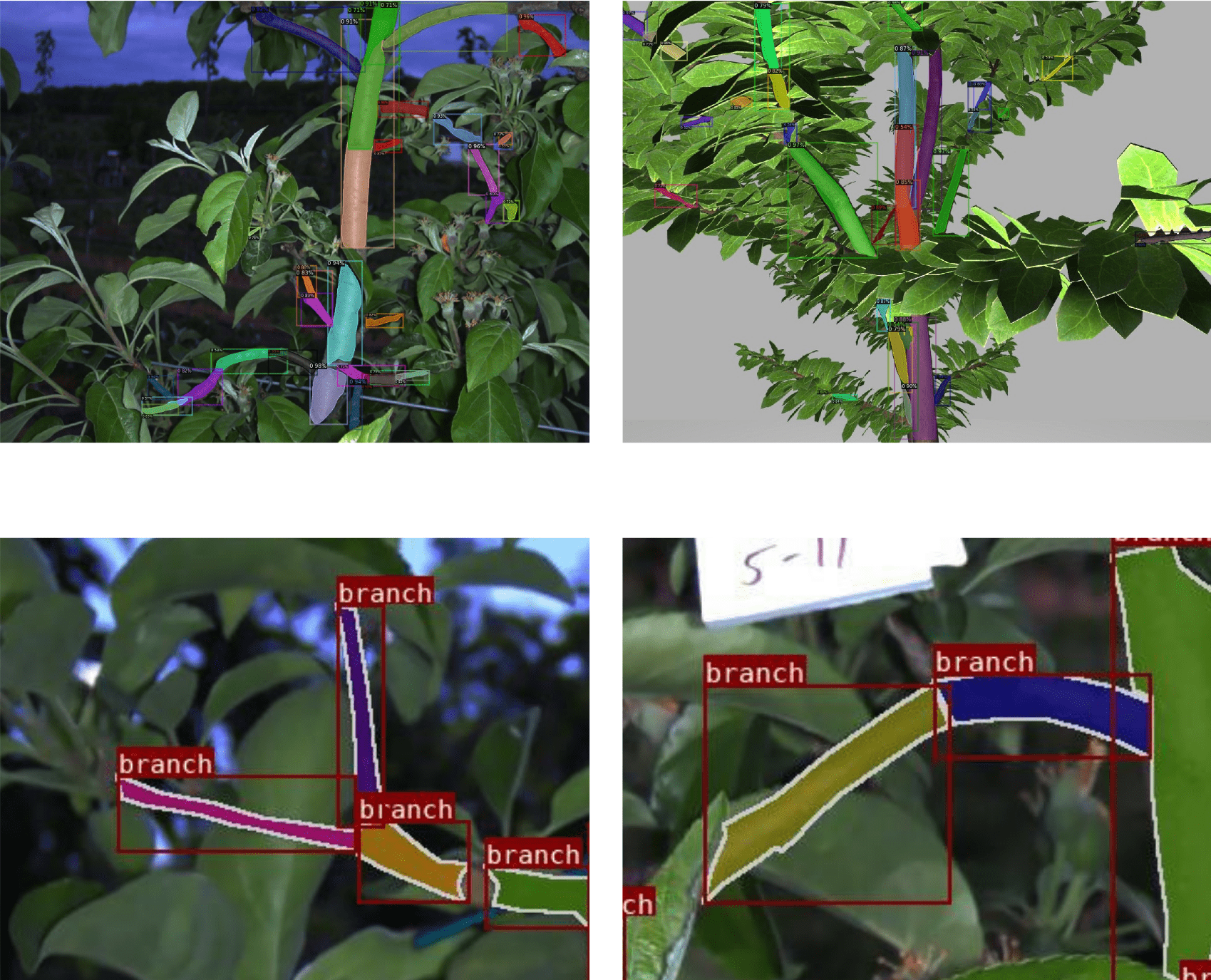}
\setlength{\unitlength}{1cm}
\begin{picture}(0,0)
\put(-3.9, 2.8){\footnotesize (a)} 
\put(-3.9, -0.3){\footnotesize (b)} 
\end{picture}
\caption{(a) Instance segmentation results on a real tree image (left) and a synthetic tree image (right). (b) Tree forks (left) or angled branch instances (right) are labeled as multiple branch instances in the training dataset. }
\label{fig:instance_segmentation} 
\vspace{-15pt}
\end{figure}
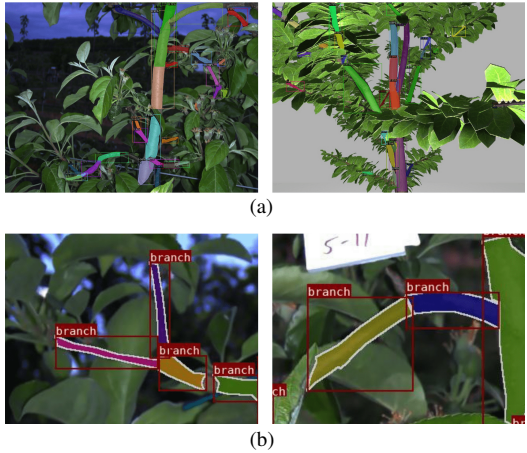

Our branch segmentation is based on the Mask R-CNN \cite{he2017mask} instance segmentation network with a Feature Pyramid Network and ResNet50 backbone. The model takes as input $1440\times1080$ images and outputs instance segmentation masks as well as its confidence scores ranging from 0 to 1. The network was trained on 130 manually labeled images (105 real images and 25 synthetic images) for 2000 iterations on an NVIDIA GeForce RTX 3070. Sample segmentation results are depicted in Fig.~\ref{fig:instance_segmentation}(a). In order to perform line-fitting on the point clusters (further described in Sec.~\ref{subsec:likelihood_map}), we deliberately labeled each branch instances in the training images to be a slender polygon such that the projected branch point cloud cluster in 3D space is also piece-wise slender. For example, a tree fork or a branch instance with a significant change in direction is labeled as two or more branch instances instead of a single instance as shown in Fig.~\ref{fig:instance_segmentation}(b).

\subsection{Skeleton Occupancy Likelihood Map}
\label{subsec:likelihood_map}

We now propose our novel approach to hypothesize on the presence of occluded branches by approximating a 3D likelihood map in the form of an occupancy grid, where the $i$-th grid voxel $\mathbf{m}_i$ stores the skeleton occupancy probability $p_\ell(\textbf{m}_i)\in[0,1]$. 
The approximation is driven by a heuristic assumption: Given a visible branch instance, it is likely that the branch structure extends longer along its growth direction based on the observation that branches in nature generally grow straight. We model this as individually observed probability distributions $p_o$ with an ellipsoidal contour obtained from the branch point cloud clusters as follows.

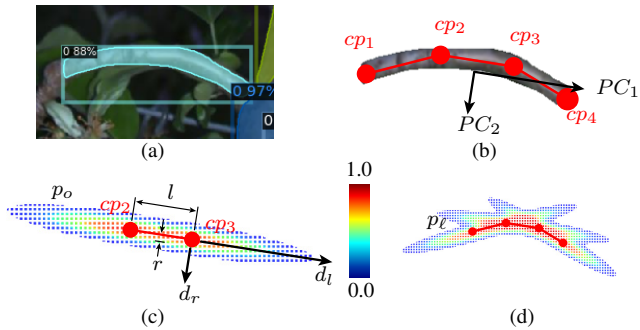
\begin{figure}[h!]
\centering 
\includegraphics[width=1.0\columnwidth]{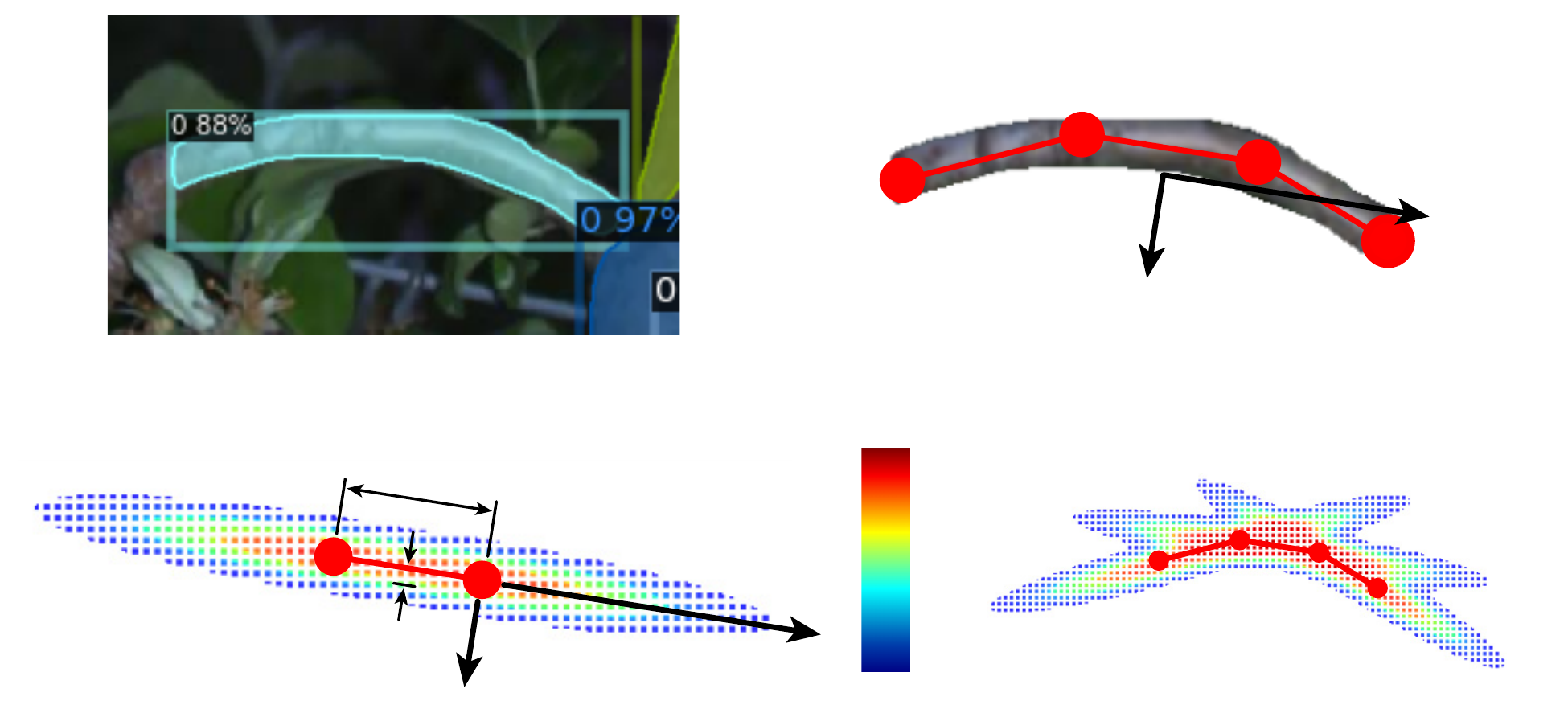}
\vspace*{-2mm}
\setlength{\unitlength}{1cm}
\begin{picture}(0,0)
\put(-2.4, 2.2){\footnotesize (a)} 
\put(2.0, 2.2){\footnotesize (b)} 
\put(-2.4, 0.0){\footnotesize (c)} 
\put(2.5, 0.0){\footnotesize (d)} 
\put(3.65, 3.05){\footnotesize $PC_1$} 
\put(1.8, 2.55){\footnotesize $PC_2$} 
\put(0.3, 3.7){\footnotesize \changed{$cp_1$}} 
\put(1.5, 3.9){\footnotesize \changed{$cp_2$}} 
\put(2.5, 3.75){\footnotesize \changed{$cp_3$}} 
\put(3.25, 2.7){\footnotesize \changed{$cp_4$}} 

\put(0.32, 1.95){\footnotesize 1.0} 
\put(0.35, 0.35){\footnotesize 0.0} 
\put(1.4, 1.3){\footnotesize $p_\ell$} 
\put(-3.59, 1.7){\footnotesize $p_o$} 
\put(-2.95, 1.5){\footnotesize \changed{$cp_2$}} 
\put(-1.55, 1.25){\footnotesize \changed{$cp_3$}} 
\put(-0.09, 0.55){\footnotesize $d_l$} 
\put(-1.9, 0.3){\footnotesize $d_r$} 
\put(-2.05, 1.65){\footnotesize $l$} 
\put(-2.25, 0.7){\footnotesize $r$} 

\end{picture}
\caption{(a) A single instance of a segmented branch component with a confidence score of 0.88. (b) The branch instance is projected into 3D space as a dense point cloud cluster. We compute the principal components ($PC_i$) and fit a B-spline curve with four control points ($cp_i$) on the point cloud cluster. (c) $p_o$ is obtained from each line segment, (d) which updates $p_\ell$ using equation~(\ref{eq:independent_events}).}
\label{fig:likelihood_map} 
\vspace{-5pt}
\end{figure}

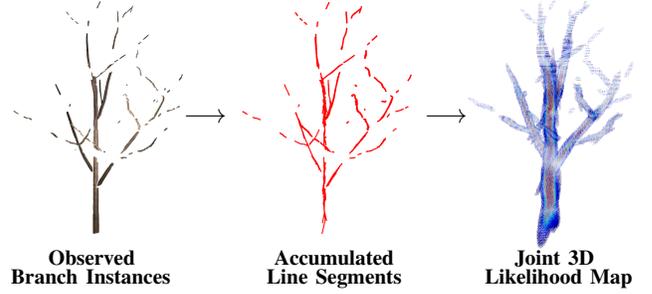
\begin{figure}[h!]
\centering 
\includegraphics[width=1.0\columnwidth]{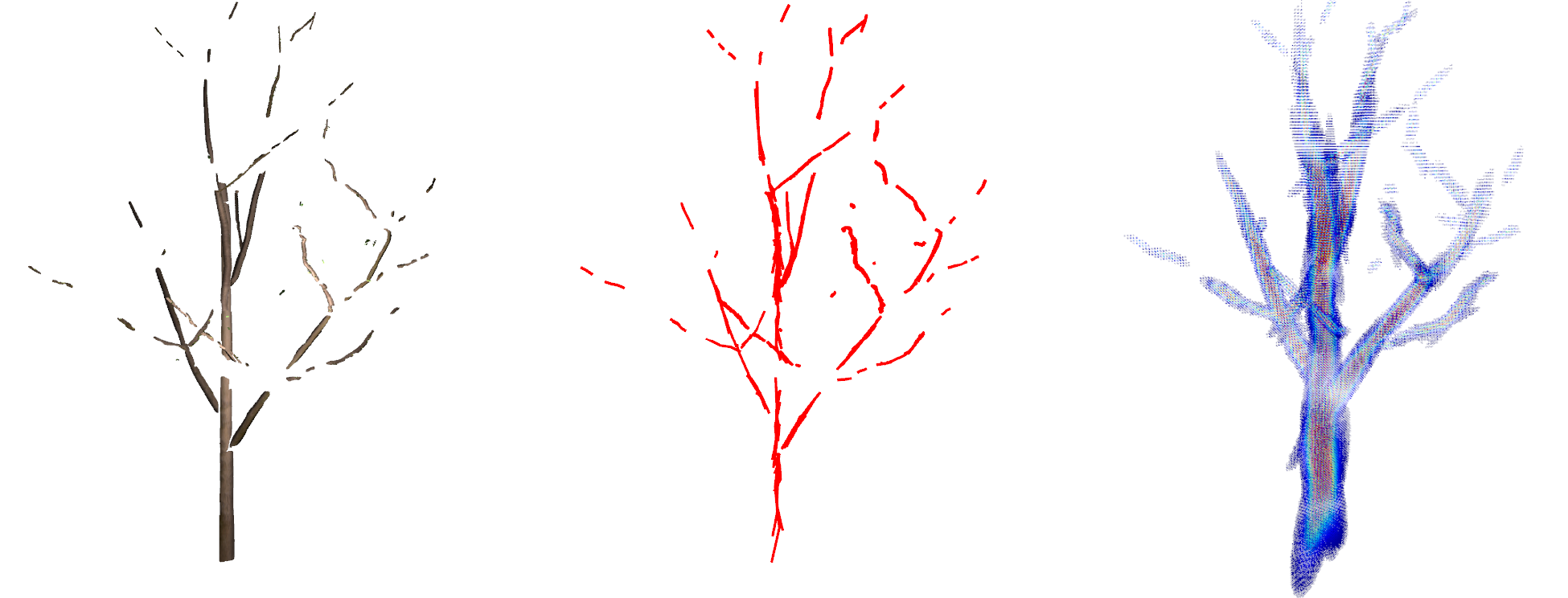}
\setlength{\unitlength}{1cm}
\begin{picture}(0,0)
\put(-3.7, 0.2){\footnotesize \textbf{Observed}} 
\put(-4.2, -0.05){\footnotesize \textbf{Branch Instances}} 
\put(-0.7, 0.2){\footnotesize \textbf{Accumulated}} 
\put(-0.8, -0.05){\footnotesize \textbf{Line Segments}} 
\put(2.5, 0.2){\footnotesize \textbf{Joint 3D}} 
\put(2.1, -0.05){\footnotesize \textbf{Likelihood Map}} 
\put(-1.9, 2.1){$\longrightarrow$} 
\put(1.3, 2.1){$\longrightarrow$} 
\end{picture}
\caption{Visualization of the overall joint 3D likelihood map (right) obtained from accumulated line segments (middle), which were extracted from observed branch clusters over the sequence of images (left).}
\label{fig:joint_likelihood_map} 
\vspace{-13pt}
\end{figure}

%The probability distribution of the joint 3D likelihood map $p_\ell$ is updated by individually observed probability distributions $p_o$ obtained from each branch point cloud cluster as follows.
A 3-dimensional B-Spline curve of degree-one with four control points is approximated on a point cloud cluster using the least squares method \cite{piegl1996nurbs}. This results in three connected line segments that best represent the skeletal geometry of the point cloud cluster as shown in Fig.~\ref{fig:likelihood_map}(b), which are accumulated over the sequence of images (Fig.~\ref{fig:joint_likelihood_map}). The radius of the branch instance is also estimated by performing principal component analysis on the point cloud cluster (see Fig.~\ref{fig:likelihood_map}(b)), where the distance between the maxima and minima points along the second principal component is approximated to be the diameter of the branch instance. %To preserve centeredness of the tree skeleton, edges are translated along the direction of the camera center to the cluster center by the radius.
The probability distribution of the joint likelihood map $p_\ell$ is updated by $p_o$ based on the following rules:
%The probability distribution of voxel occupancy $p_o$ imposed by each observed edge is assigned based on the following rules:
\begin{enumerate}
    \item $p_o$ of the voxels containing the line segment is equal to the mask confidence score obtained from the instance segmentation network.
    \item $p_o$ of the voxels in the proximity of the line segment diminishes in the axial and radial directions along an ellipsoidal contour (Fig.~\ref{fig:likelihood_map}(c)):
    %at a rate inversely proportional to the edge length and estimated radius, respectively (Fig.~\ref{fig:likelihood_map}(c)).
    \begin{equation}
        p_o=c-\frac{c}{k}\sqrt{\left(\frac{d_l}{l}\right)^2+\left(\frac{d_r}{r}\right)^2}
    \end{equation}
    where $c$ is the mask confidence score, $l$ is the line segment length, and $r$ is the estimated radius. $d_l$ and $d_r$ is the axial and radial distance of the voxel from the line segment, and $k$ is a parameter that controls the rate at which $p_o$ decreases as $d_l$ and $d_r$ increases. 
    For example, a large $k$ results in a small decreasing rate, effectively enlarging the size of the ellipsoid. We empirically set $k=3$ in our experiments.
\item The voxel $\mathbf{m}_i$ of the 3D likelihood map is updated by the observed occupancy probability $p_o$ according to the following equation (Fig.~\ref{fig:likelihood_map}(d)):
    \begin{equation}
    p_\ell(\mathbf{m}_i)=1-[1-p_\ell(\mathbf{m}_i)][1-p_o(\mathbf{m}_i)]
    \label{eq:independent_events}
    \end{equation}
\end{enumerate}
%\todo{Describe the assumptions that allows us to use the update equation.}
The resulting joint likelihood map updated by $p_o$ from all line segments %observed branch clusters 
is depicted in Fig.~\ref{fig:joint_likelihood_map}. The 3D likelihood map effectively has higher occupancy probability in voxels that are directly intersected by a line segment,
as well as in voxels that are jointly influenced by more than one line segment.
%where a line segment directly intersect,  

\subsection{Skeleton Extraction}
\label{subsec:skeletonization}

The final tree skeleton $G_T=(V_T, E_T)$ is an undirected acyclic graph with vertices $V_T$ and edges $E_T$ representing the topological structure of the tree canopy. This is obtained by post-processing the accumulated line segments with the joint likelihood map as follows. 

We first obtain an initial tree skeleton $G_\text{init}=(V_\text{init}, E_\text{init})$ by consolidating the accumulated line segments. To do so, all line segments are converted into a set of vertices by sampling points equally spaced along the line (in practice, we set the spacing to be the voxel size of the likelihood map). 
%The vertices are consolidated into a skeletal curve using a modified Laplacian smoothing method which reposition each vertex to the mean of its neighbors within a search radius:
The vertices are consolidated into a skeletal curve using the Laplacian smoothing method which repositions each vertex to the mean of its \textit{k}-nearest neighbors:
\begin{equation}
    v_i = \frac{1}{k}\sum_{j=1}^N v_j
\end{equation}
where $k$ is the number of neighborhood vertices, $v_j$ is the position of neighbor vertex $j$, and $v_i$ is the new position of vertex $i$. %The search radius for each point is dynamically computed by setting it to be proportional to the shape factor (aspect ratio) of \textit{k}-nearest neighbor points, such that neighborhood clusters with high (low) aspect ratio results in a large (small) search radius.
Laplacian smoothing is iteratively applied until the total change in vertex position per iteration converges below a threshold, which results in a set of well-refined skeletal vertices $V_\text{init}$. The edges $E_\text{init}$ are initialized by building a Euclidean minimum spanning tree from $V_\text{init}$ with edges that are longer than the voxel size removed.
%in a well refined set of skeletal vertices of observed branch segments. 
%The initial tree skeleton is finally obtained by building a Euclidean minimum spanning tree from the obtained skeleton vertices \todo{with radius threshold}.  

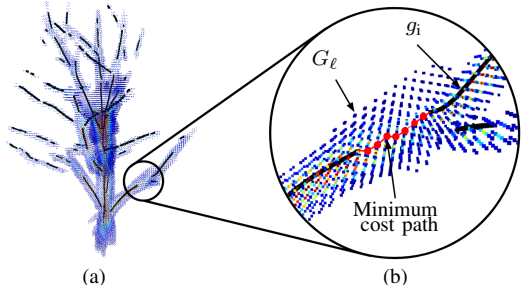
\begin{figure}[h!]
\centering 
\includegraphics[width=0.8\columnwidth]{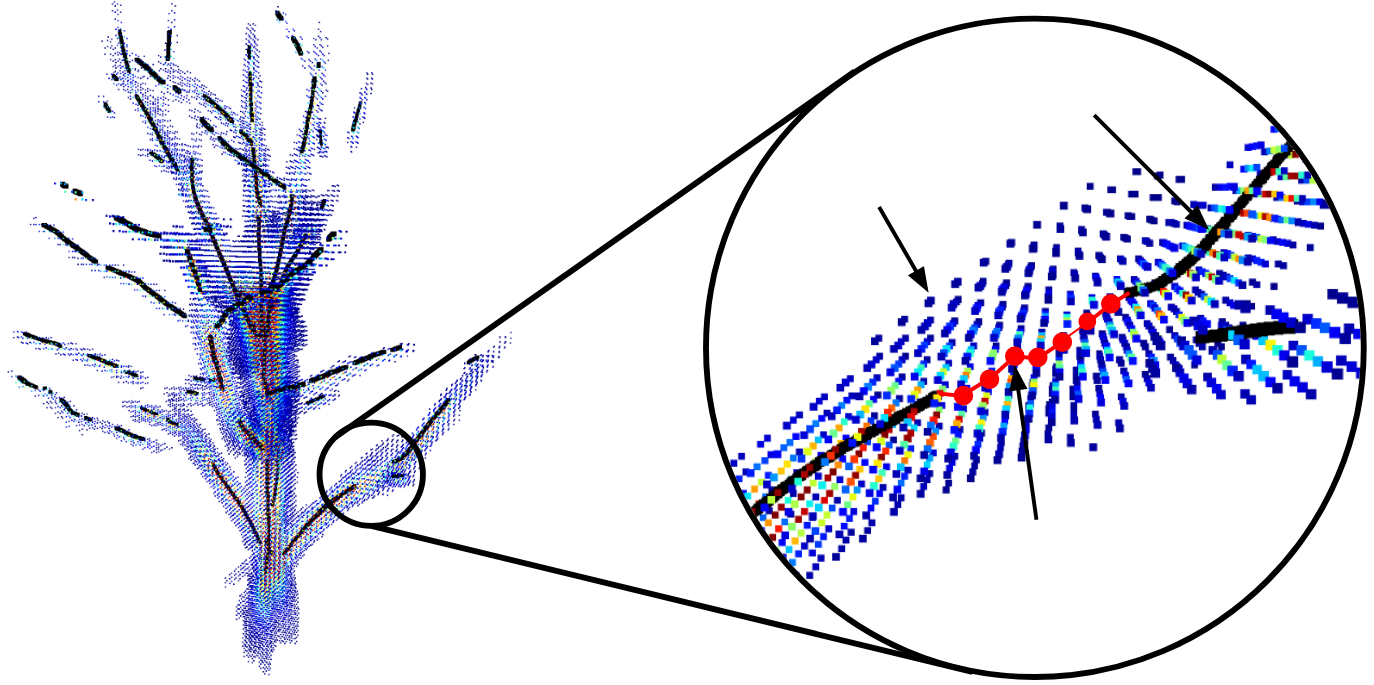}
\setlength{\unitlength}{1cm}
\begin{picture}(0,0)
\put(-6.1, -0.3){\footnotesize (a)}
\put(-2.1, -0.3){\footnotesize (b)}
\put(-1.8, 3.05){\footnotesize $g_\text{i}$} 
\put(-2.5, 0.6){\footnotesize Minimum} 
\put(-2.4, 0.4){\footnotesize cost path} 
\put(-3.05, 2.6){\footnotesize $G_\ell$} 
\end{picture}
\caption{(a) The initial skeleton $G_\text{init}$ shown as the curves colored black overlapped with the weighted likelihood graph $G_\ell$. $G_\text{init}$ is composed of disjoint subgraphs $g_i$ due to occlusion. (b) We search for minimum cost paths in $G_\ell$ to join disconnected subgraphs $g_i$. }
\label{fig:min_cost_path} 
\vspace{-5pt}
\end{figure}

Due to occlusion by leaves and branches, $G_\text{init}=\{g_i\}$ is composed of disjoint skeletal subgraphs $g_i$ of a single tree. 
%Due to occlusion by leaves and branches, the initial skeleton are disconnected skeletal subgraphs of visible branch structures. In order to predict the occluded parts of the tree, we join maximally connected subgraphs of the initial skeleton by searching for minimum cost paths in the 3D likelihood map as shown in Algorithm 1. 
In order to predict the occluded parts of the tree, we join maximally connected subgraphs in $G_\text{init}$ by searching for minimum cost paths in the 3D likelihood map as illustrated in Fig.~\ref{fig:min_cost_path} and summarized in Algorithm 1. 
The likelihood map is first converted into a weighted graph $G_\ell$ by adding undirected edges between all $3\times3\times3$ adjacent nodes (Line 2, Algorithm 1). The cost of an edge connecting vertex $u$ and $v$ is set to be the negative log of the average occupancy probability: 
\begin{equation}
    c(e_{uv})=-\log\left(\frac{p_\ell(\mathbf{m}_u)+p_\ell(\mathbf{m}_v)}{2}\right)
\end{equation}
which results in low cost for edges between nodes with high occupancy probability. 
%Using the weighted likelihood graph $G_\ell$, the subgraphs in $G_\text{init}$ are iteratively connected by searching for minimum cost paths using Dijkstra's algorithm until no more paths exists.
Using the weighted likelihood graph $G_\ell$, the minimum cost path between all pairwise subgraph combinations in $G_T$ (initialized with $G_\text{init}$) are computed (Line 5-7, Algorithm 1). The path with the minimum cost is added to $G_T$ (Line 9-10, Algorithm 1). This process is repeated until no more paths are found resulting in our final skeleton output $G_T$, composed of skeletal curves that were directly observed, as well as predicted skeletal curves that were unobservable due to occlusion. 

\begin{table}[h]
\centering
\small
\begin{tabularx}{\columnwidth}{l}
%\hline
%\Xhline{2\arrayrulewidth}
\toprule
\makecell[cl]{\textbf{Algorithm 1} Likelihood Map Path Search} \\ %\hline 
\midrule
\makecell[cl]{
\textbf{Input:} Initial Skeleton $G_\text{init}$, Likelihood Map $p_\ell$\\
\textbf{Output:} Tree Skeleton $G_T$\\
{\footnotesize 1:} \hspace{1mm} $G_T\leftarrow G_\text{init}$\\
{\footnotesize 2:} \hspace{1mm} $G_\ell\leftarrow\texttt{WeightedUndirectedGraph}(p_\ell)$\\
{\footnotesize 3:} \hspace{1mm} \textbf{Repeat}\\
{\footnotesize 4:} \hspace{5mm} $\mathcal{P}\leftarrow\texttt{EmptyArray}()$\\
{\footnotesize 5:} \hspace{5mm} \textbf{for all} pairwise subgraph combination $(g_u, g_v)\in G_T$ \textbf{do}\\
{\footnotesize 6:} \hspace{9mm} $\rho \leftarrow\texttt{DijkstrasMinPath}(g_u, g_v, G_\ell)$\\
{\footnotesize 7:} \hspace{9mm} $\mathcal{P}.\texttt{append}(\rho)$\\
{\footnotesize 8:} \hspace{5mm} \textbf{end for}\\
{\footnotesize 9:} \hspace{5mm} $\rho_\text{min}\leftarrow\texttt{MinCostPath}(\mathcal{P})$\\
{\footnotesize 10:} \hspace{3.2mm} $G_T\leftarrow \texttt{JoinSubgraphs}(G_T, \rho_\text{min})$\\
{\footnotesize 11:} \hspace{0mm} \textbf{Until} no more paths are found\\
} \\ 
\bottomrule
\end{tabularx}
%\vspace{-5pt}
\end{table}

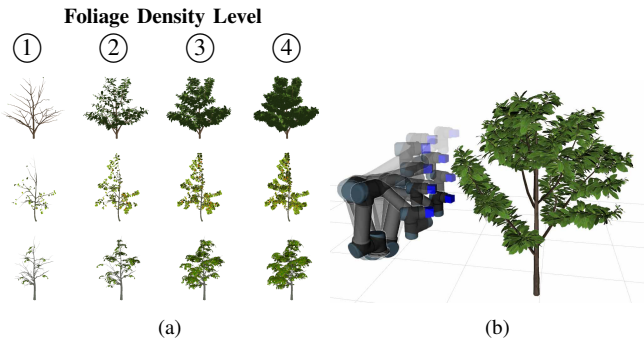
\begin{figure}[h]
\centering 
\includegraphics[width=\columnwidth]{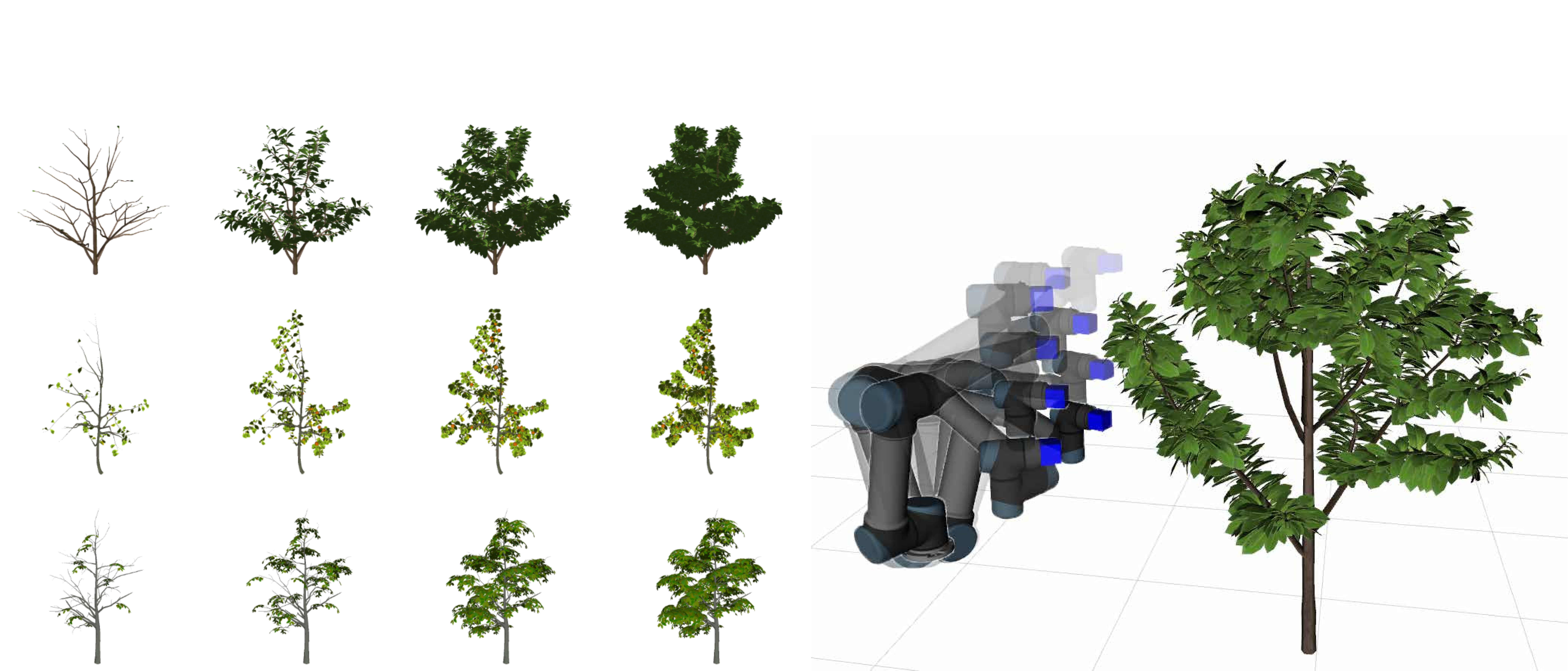}
\vspace*{-2mm}
\setlength{\unitlength}{1cm}
\begin{picture}(0,0)
\put(-2.2, 0.0){\footnotesize (a)} 
\put(2.15, 0.0){\footnotesize (b)} 
\put(-3.45, 4.15){\footnotesize \textbf{Foliage Density Level}} 
\put(-4.15, 3.7){\footnotesize {\large \textcircled{\small 1}}} 
\put(-3.0, 3.7){\footnotesize {\large \textcircled{\small 2}}} 
\put(-1.85, 3.7){\footnotesize {\large \textcircled{\small 3}}} 
\put(-0.70, 3.7){\footnotesize {\large \textcircled{\small 4}}} 

\end{picture}
\caption{(a) Synthetic meshes of an oak tree (top row), apple tree (middle row), and a walnut tree (bottom row) with varying foliage density increasing from left to right. Trees in the same row have the same structural topology. Our synthetic dataset consists of ten unique topology per tree species. (b) The robot is controlled to take images of the tree canopy from 10 different viewpoints in the Gazebo simulation environment.}
\label{fig:synthetic_tree_dataset} 
\vspace{-10pt}
\end{figure}

\section{Experiments}
\label{sec:experiments}

We evaluate our proposed tree skeletonization method through quantitative evaluation and visual assessment. For the former, we collected a synthetic tree mesh dataset with known ground truth skeleton and propose metrics to measure the precision of the skeleton as well as the effectiveness of predicting unseen branch skeletons. Visual assessment is presented for the simulated dataset as well as a real tree dataset collected from an apple orchard.

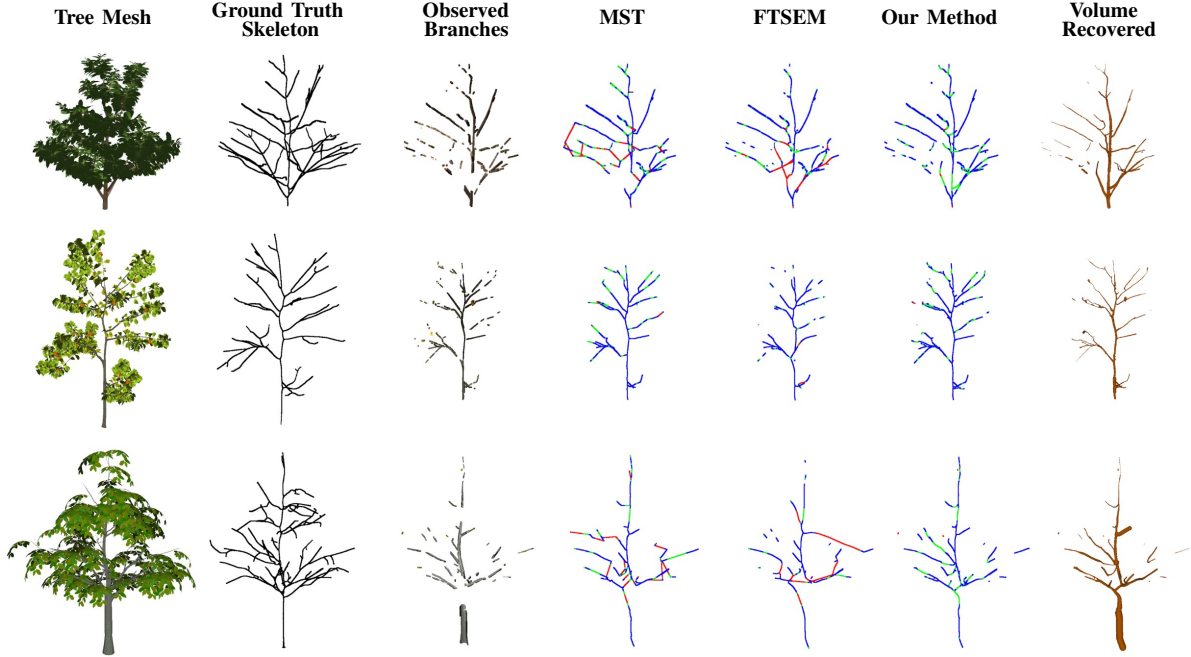
\begin{figure*}[h!]
\centering 
\includegraphics[width=0.89\textwidth]{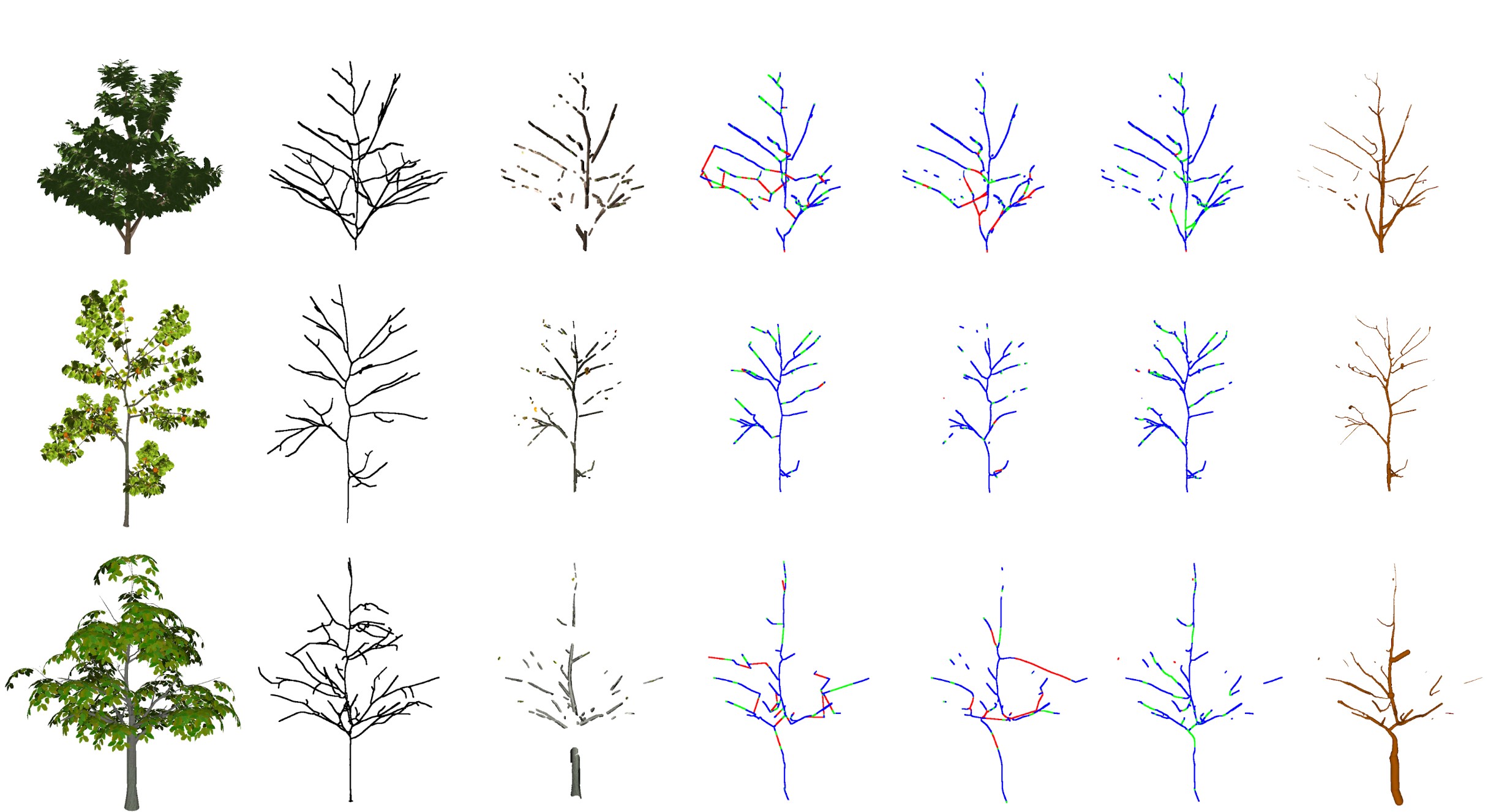}
\vspace*{-3mm}
\setlength{\unitlength}{1cm}
\begin{picture}(0,0)
\put(-15.4, 8.4){\footnotesize \textbf{Tree Mesh}} 
\put(-13.3, 8.5){\footnotesize \textbf{Ground Truth}} 
\put(-12.9, 8.25){\footnotesize \textbf{Skeleton}} 
\put(-10.5, 8.5){\footnotesize \textbf{Observed}} 
\put(-10.48, 8.25){\footnotesize \textbf{Branches}} 
\put(-8.15, 8.4){\footnotesize \textbf{MST}} 
\put(-6.1, 8.4){\footnotesize \textbf{FTSEM}} 
\put(-4.4, 8.4){\footnotesize \textbf{Our Method}} 
\put(-1.9, 8.5){\footnotesize \textbf{Volume}} 
\put(-2.0, 8.25){\footnotesize \textbf{Recovered}} 
\end{picture}
\caption{Experiment results of synthetic trees including oak (top row), apple (middle row), and walnut (bottom row) trees with foliage density level \textcircled{\scriptsize 3}. We compare our method against MST-based and FTSEM-based methods. The extracted skeleton is color coded: $TP$, $TP_\text{occ}$, and $FP$ vertices are colored blue, green, and red, respectively. The right-most column depicts the skeleton extracted from our method with recovered volume based on the radius information of the skeletal vertices.}
\label{fig:synthetic_tree_results} 
\vspace{-15pt}
\end{figure*}

\subsection{Experiments on Synthetic Trees}

Experiments in simulation were performed on three different types of synthetic trees including oak, apple, and walnut trees. For each species, ten different structural topologies with four levels of varying foliage density were generated as depicted in Fig.~\ref{fig:synthetic_tree_dataset}(a), totalling to 120 trees. The synthetic tree meshes were created in SpeedTree\footnote{\url{www.speedtree.com}} which were imported into a Gazebo\footnote{\url{www.gazebosim.org}} simulation environment to model an RGB-D camera attached to a UR5 robot arm. For each tree, the robot is controlled to collect images from 10 different viewpoints as shown in Fig.~\ref{fig:synthetic_tree_dataset}(b), which are sequentially passed onto our skeletonization pipeline.  

We propose the following metrics to assess the correctness of the constructed skeleton as well as the effectiveness of our approach:
\begin{itemize}
    \item \textbf{Skeleton Precision and Recall:} For each synthetic tree, vertices of the output skeleton with a corresponding point in the ground truth skeleton within a 0.02m radius are labeled as true positives ($TP$), while vertices without correspondence are labeled as false positives ($FP$). Vertices of the ground truth skeleton without a corresponding point in the output skeleton are labeled as false negatives ($FN$).
    %\begin{equation}
    %    \text{Precision} = \frac{TP}{TP+FP}
    %\end{equation}
    \item \textbf{Occluded Skeleton Ratio (OSR):} To quantify the effectiveness of our approach with regards to predicting occluded branch skeletons correctly, we measure the percentage of true positive vertices that were obtained from the minimum cost paths ($TP_\text{occ}$) versus the total number of vertices in the constructed skeleton: 
    \begin{equation}
        \text{OSR}=\frac{TP_{occ}}{TP+FP}
    \end{equation} 
\end{itemize}

\newcolumntype{A}{>{\centering\arraybackslash\hsize=.1\hsize}X}
\newcolumntype{B}{>{\centering\arraybackslash\hsize=.1\hsize}X}
\newcolumntype{D}{>{\centering\arraybackslash\hsize=.1\hsize}X}
\newcolumntype{E}{>{\centering\arraybackslash\hsize=.1\hsize}X} 
\newcolumntype{F}{>{\centering\arraybackslash\hsize=.1\hsize}X}
\newcolumntype{G}{>{\centering\arraybackslash\hsize=.1\hsize}X}
\newcolumntype{H}{>{\centering\arraybackslash\hsize=.1\hsize}X}
\newcolumntype{I}{>{\centering\arraybackslash\hsize=.1\hsize}X}
\newcolumntype{J}{>{\centering\arraybackslash\hsize=.1\hsize}X}
\newcolumntype{K}{>{\centering\arraybackslash\hsize=.1\hsize}X}
\newcolumntype{L}{>{\centering\arraybackslash\hsize=.1\hsize}X}
\begin{table}
\caption{Experiment results of tree skeletonization\\on synthetic trees.}
\vspace{-8pt}
\centering 
\begin{tabularx}{\columnwidth}{@{}ABDEFGHIJKL@{}}  %{@{}gddnnuogu@{}} 
\hline\hline %inserts double horizontal lines
%\rule{0pt}{2.5ex}
\multirow{2}{*}{{\shortstack[c]{\vspace{1pt}\\Tree\\Type}}} & \multirow{2}{*}{{\shortstack[c]{\vspace{5pt}\\FD}}} & \multicolumn{3}{p{1.8cm}}{\centering MST} & \multicolumn{3}{p{1.8cm}}{\centering FTSEM} & \multicolumn{3}{p{1.8cm}}{\centering Ours}\\
\cmidrule(lr){3-5} \cmidrule(lr){6-8} \cmidrule(lr){9-11}
%\rule{0pt}{2.5ex}
&  &  P & R & OSR &  P & R & OSR &  P & R & OSR \\[0.5ex] % inserts table
%\rule{0 pt}{2.5ex}
\hline 
\multirow{4}{*}{{\shortstack[c]{\vspace{3pt} \\Oak}}} 
 & 1 &  \textbf{0.99} & 0.88 & \textbf{0.05} & 0.97 & 0.86 & 0.03 & \textbf{0.99} & \textbf{0.91} & 0.04\\[0.5ex]
 & 2 &  0.96 & 0.78 & \textbf{0.08} & 0.93 & 0.73  & 0.04 & \textbf{0.98} & \textbf{0.82} & \textbf{0.08}\\[0.5ex]
 & 3 &  0.89 & 0.51 & 0.14 & 0.88 & 0.46 & 0.11 & \textbf{0.96} & \textbf{0.54} & \textbf{0.15}\\[0.5ex]
 & 4 &  0.86 & 0.29 & 0.17 & 0.90 & 0.27 & 0.16 & \textbf{0.98} & \textbf{0.32} & \textbf{0.19}\\[0.5ex]
\hline 
 \multirow{4}{*}{{\shortstack[c]{\vspace{3pt} \\Apple}}} 
 & 1 &  0.94 & 0.71 & \textbf{0.14} & 0.91 & 0.64 & 0.07 & \textbf{0.98} & \textbf{0.73} & 0.12\\[0.5ex]
 & 2 &  0.95 & 0.62 & \textbf{0.16} & 0.94 & 0.51 & 0.05 & \textbf{0.99} & \textbf{0.65} & 0.13\\[0.5ex]
 & 3 &  0.95 & 0.49 & \textbf{0.14} & 0.91 & 0.40 & 0.05 & \textbf{0.99} & \textbf{0.52} & \textbf{0.14}\\[0.5ex]
 & 4 &  0.94 & 0.39 & 0.15 & 0.91 & 0.32 & 0.06 & \textbf{0.97} & \textbf{0.42} & \textbf{0.16}\\[0.5ex]
\hline 
 \multirow{4}{*}{{\shortstack[c]{\vspace{3pt} \\Walnut}}} 
 & 1 &  0.95 & 0.70 & \textbf{0.10} & 0.95 & 0.65 & 0.07 & \textbf{0.99} & \textbf{0.72} & 0.08\\[0.5ex]
 & 2 &  0.93 & 0.64 & \textbf{0.12} & 0.93 & 0.56 & 0.06 & \textbf{0.98} & \textbf{0.66} & 0.10\\[0.5ex]
 & 3 &  0.86 & 0.48 & 0.15 & 0.84 & 0.39 & 0.10 & \textbf{0.96} & \textbf{0.49} & \textbf{0.16}\\[0.5ex]
 & 4 &  0.80 & \textbf{0.34} & 0.16 & 0.84 & 0.29 & 0.13 & \textbf{0.97} & \textbf{0.34} & \textbf{0.17}\\[0.5ex]
\hline\hline
\end{tabularx}
\begin{flushleft}
\textit{* \textbf{FD}: Foliage Density, \textbf{P}: Precision, \textbf{R}: Recall }
\end{flushleft}
\vspace{-12pt}
\begin{flushleft}
\textit{* Each score is averaged over 10 trees belonging to the tree type and foliage density.}
\end{flushleft}
\label{table:results} 
\vspace{-25pt}
\end{table}

\begin{figure*}[h]
\centering 
\includegraphics[width=\textwidth]{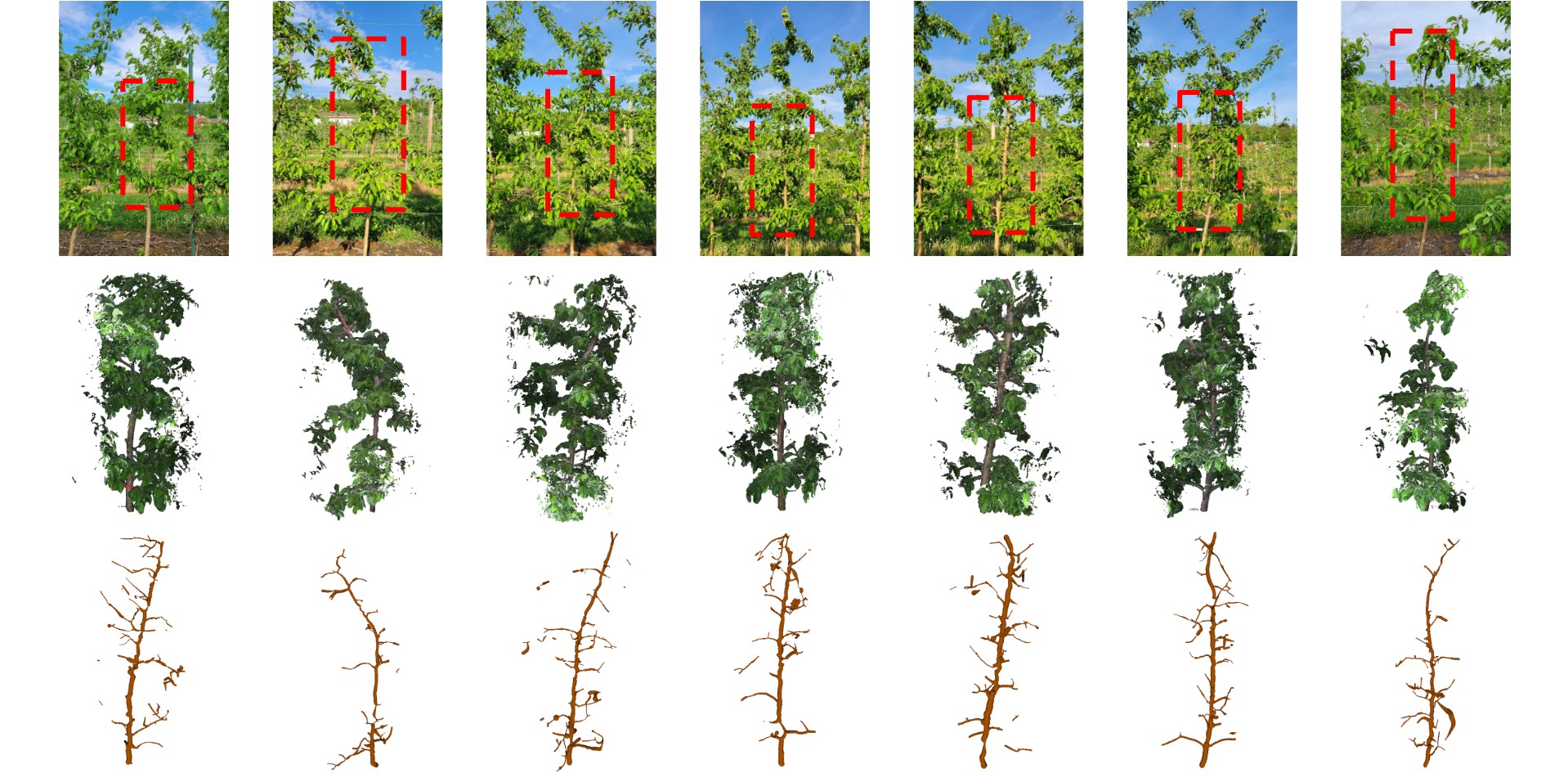}
\vspace*{-7mm}
\setlength{\unitlength}{1cm}
\begin{picture}(0,0)
%\put(-0.2, 7.3){\footnotesize (a)} 
%\put(-0.2, 3.7){\footnotesize (b)} 
%\put(-0.2, 0.0){\footnotesize (c)} 
\end{picture}
\caption{Experiment results of seven real apple trees. The first row shows the images of the apple tree. The region bounded by the red dashed square is imaged by the robot with a stereo camera, corresponding to the 3D reconstructed point cloud depicted in the second row. The result of our skeletonization method is shown in the third row with recovered volume by replacing skeleton vertices with spheres of corresponding radius.}
\label{fig:real_tree_results} 
\vspace{-10pt}
\end{figure*}

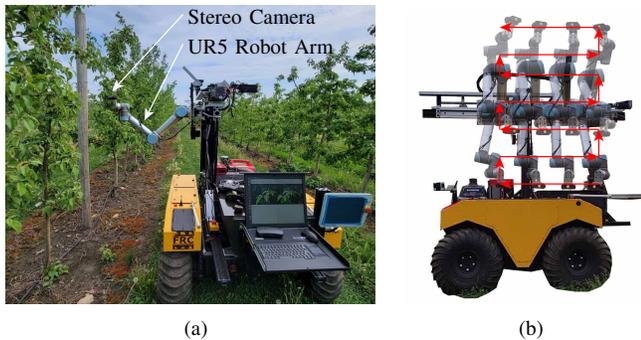
\begin{figure}[h]
\centering 
\includegraphics[width=\columnwidth]{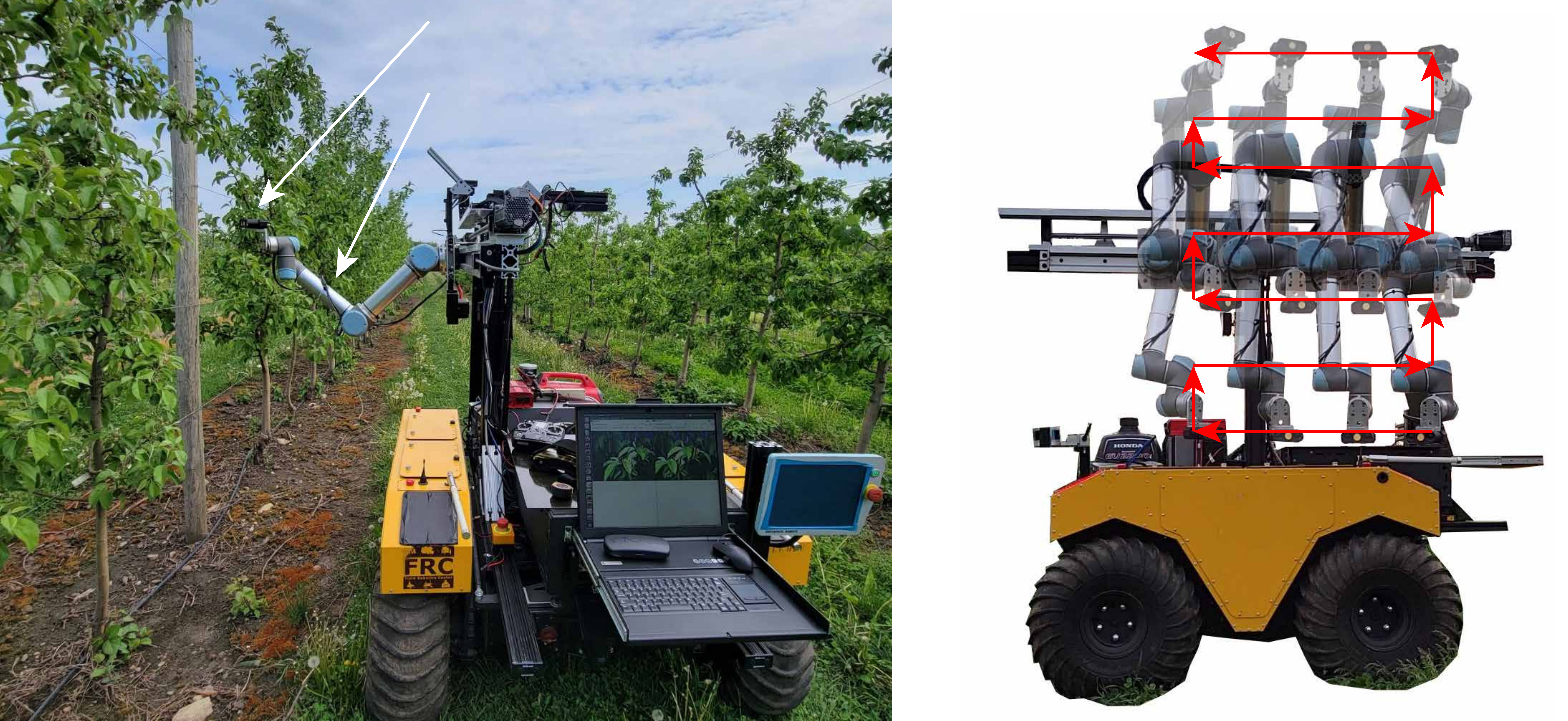}
\vspace*{-3mm}
\setlength{\unitlength}{1cm}
\begin{picture}(0,0)
\put(-2.0, 0.0){\footnotesize (a)} 
\put(2.45, 0.0){\footnotesize (b)} 
\put(-1.95, 4.15){\footnotesize Stereo Camera} 
\put(-1.95, 3.75){\footnotesize UR5 Robot Arm} 
\end{picture}
\caption{(a) Our hardware setup for collecting real world data at the UMass Cold Spring Orchard, featuring a stereo camera attached to the UR5 robot arm. (b) The robot motion path is shown by the red arrows, where it collects 70 stereo image pairs of each tree canopy at equally distributed waypoints throughout the motion path. }
\label{fig:bumblebee} 
\vspace{-15pt}
\end{figure}

We compare our results to those obtained from a minimum spanning tree (MST) based method \cite{livny2010automatic, gorte2004structuring} as well as a method presented in FTSEM \cite{sun2022fast}. For the MST-based method, our process of connecting disjoint subgraphs via searching for minimum cost paths in the likelihood map (described in Sec.~\ref{subsec:skeletonization})  is replaced by returning the Euclidean minimum spanning tree built from $V_\text{init}$ as the final skeleton. For FTSEM, we use the breakpoint connection method \cite{sun2022fast} to connect disjoint subgraphs by checking for distance and angle conditions between vectors computed from edges in subgraph pairs. Since the code for FTSEM is not open sourced,  we implement it ourselves based on the available details.

The results are summarized in Table~\ref{table:results} and visualized in Fig.~\ref{fig:synthetic_tree_results}. The precision, recall, and OSR of our method averaged over all 120 synthetic trees is 0.98, 0.59 and 0.11, respectively.
Our method outperforms both baselines in terms of precision and recall as it joins disconnected skeletons only when a path in the likelihood map exists. In contrast, the MST-based method greedily joins all disconnected skeletons. Although this results in higher OSR in less occluded scenes (Table~\ref{table:results}, foliage density 1 \& 2), the relative performance of MST deteriorates with increasing foliage density (Table~\ref{table:results}, foliage density 3 \& 4). As a result, our method outperforms MST in terms of precision, recall and OSR where there is high occlusion.

\subsection{Experiments on Real Trees}
Experiments on real trees were performed on a dataset that consists of 7 apple trees  (Fig.~\ref{fig:real_tree_results}) collected from the University of Massachusetts Amherst Cold Spring Orchard. For each tree, stereo images were collected from 70 different viewpoints using a flash stereo camera \cite{9636542} attached to the UR5 robot arm (Fig.~\ref{fig:bumblebee}(a)). The robot was controlled to follow a motion path as depicted in Fig.~\ref{fig:bumblebee}(b), where the viewpoints are set to be at equally distributed waypoints throughout the motion path. Due to the robot's limited workspace, we were able to capture a range of approximately 1.5 meters in height for each tree canopy. 

Our results are visualized in Fig.~\ref{fig:real_tree_results}. The OSR averaged over all seven trees is 0.14, consistent with the results obtained from the synthetic dataset. As it is difficult to obtain the ground truth skeleton for real trees, we visually compare the extracted skeleton with the 3D reconstructed tree using known camera poses from the robot. Despite considerable occlusion and noise evident in the reconstructed point cloud, the extracted skeleton is topologically correct and shows good correspondence. 

The quality of the skeleton obtained from our pipeline is contingent on the sufficiency and correctness of the detected branches. We expect that a control policy to collect images from optimal viewpoints \cite{zeng2022deep} (rather than fixed viewpoints as in our experiments) to perceive sufficient amount of branches will further improve our proposed skeletonization pipeline.

\section{Conclusion}

Our tree skeletonization method outperforms the baselines in situations with highly occluded canopies by accurately estimating unobserved skeletons. As future work, we plan to improve the algorithm runtime in addition to conducting a more rigorous evaluation on real tree datasets. Other possible directions for future work include next-best viewpoint optimization to increase the information of occluded regions in the tree canopy.
We are also interested in estimating the dynamics of the tree in response to external forces to plan for contact interactions. Ultimately, the digitized model of a tree crop in the form of a skeleton presents promising direction to developing safe and robust agricultural robotic manipulation. 

\section*{Acknowledgment} 
This work was supported in part by NSF/USDA-NIFA Cyber Physical Systems 2020-67021-31531, NSF Robust Intelligence 1956163, and NSF/USDA-NIFA AIIRA AI Research Institute 2021-67021-35329.
The authors would like to thank Daniel Cooley, Paul O'Connor, Jon Clements, Harry Freeman and Abhisesh Silwal for their help in field data collection at the University of Massachusetts Amherst Cold Spring Orchard, and Hung-Jui Huang for helpful discussions.

\FloatBarrier
\bibliographystyle{ieeetr}
\bibliography{main}

\end{document}